\documentclass[journal]{IEEEtran}
\usepackage{amsmath,amsfonts}
\usepackage{algorithmic}
\usepackage{algorithm}
\usepackage{array}
\usepackage[caption=false,font=normalsize,labelfont=sf,textfont=sf]{subfig}
\usepackage{textcomp}
\usepackage{stfloats}
\usepackage{url}
\usepackage{verbatim}
\usepackage{graphicx}
\usepackage{cite}
\usepackage[separate-uncertainty=true]{siunitx}

\usepackage{balance}
\usepackage{comment}
\usepackage{siunitx}

\cleardoublepage

\usepackage{xcolor}
\usepackage{gensymb}
\usepackage{amssymb}

\def\eg{{\em {\em e.g.},\ }}
\def\ie{{\em i.e.,\ }}

\begin{document}

\title{Multi-robot connective collaboration toward collective obstacle field traversal}

\author{{Haodi Hu, Xingjue Liao, Wuhao Du, Feifei Qian}
        % <-this % stops a space
\thanks{This work is supported by funding from the National Science Foundation (NSF) CAREER award \#2240075, %the NASA Planetary Science and Technology Through Analog Research (PSTAR) program, Award \# 80NSSC22K1313, 
and the NASA Lunar Surface Technology Research (LuSTR) program, Award \# 80NSSC24K0127. }% <-this % stops a space
%\thanks{Manuscript received April 19, 2021; revised August 16, 2021.}
}



\maketitle

\begin{abstract}
Environments with large terrain height variations present great challenges for legged robot locomotion. Drawing inspiration from fire ants' collective assembly behavior, we study strategies that can enable two ``connectable'' robots to collectively navigate over bumpy terrains with height variations larger than robot leg length. Each robot was designed to be extremely simple, with a cubical body and one rotary motor actuating four vertical peg legs that move in pairs. Two or more robots could physically connect to one another to enhance collective mobility. We performed locomotion experiments with a two-robot group, across an obstacle field filled with uniformly-distributed semi-spherical ``boulders''. Experimentally-measured robot speed suggested that the connection length between the robots has a significant effect on collective mobility: connection length $C \in [0.86, 0.9]$ robot unit body length (UBL) were able to produce sustainable movements across the obstacle field, whereas connection length $C \in [0.63, 0.84]$ and $[0.92,  1.1]$ UBL resulted in low traversability. An energy landscape based model revealed the underlying mechanism of how connection length modulated collective mobility through the system's potential energy landscape, and informed adaptation strategies for the two-robot system to adapt their connection length for traversing obstacle fields with varying spatial frequencies. Our results demonstrated that by varying the connection configuration between the robots, the two-robot system could leverage mechanical intelligence to better utilize obstacle interaction forces and produce improved locomotion. Going forward, we envision that generalized principles of robot-environment coupling can inform design and control strategies for a large group of small robots to achieve ant-like collective environment negotiation.
\end{abstract}

\begin{IEEEkeywords}
legged locomotion, rough terrain, multi-agent system 
\end{IEEEkeywords}

\section{Introduction}
%\colorbox{green}{Feifei +1} \colorbox{purple}{Haodi +3}

Inspired by animals' collective behaviors, swarm systems has been long studied in robotics~\cite{castello2019blockchain}. Traditionally, swarm research has been focused on planning algorithms to form specific patterns~\cite{kushleyev2013towards} or move to assigned locations without colliding into obstacles or one another~\cite{hettiarachchi2009distributed, guzzi2014bioinspired, preiss2017crazyswarm}, which are crucial for navigation and exploration in unknown environments~\cite{gopalakrishnan2017prvo, burgard2005coordinated}. Recent research, however, has started to shift towards incorporating physical interactions among the robots~\cite{li2019particle,ozkan2021collective}. Utilization of physical interactions can significantly boost the swarm's capabilities. %{\WD Previous sentence might need  some grammatical rework, the use of "while" is a little ambiguous. }%For example, many animals can enhance their locomotion capabilities in challenging environments by physically interacting or even connecting with one another. 
For instance, similar to fire ants that can enhance their water-repelling ability considerably by linking their bodies together to survive flood\cite{mlot2011fire},  by allowing robots to physically connect or interact with one another, they can achieve versatile and adaptive environment interactions and collectively traverse a wide variety of extreme terrains. 
%and California blackworms are able to connect their body to behave as a living material capable of mitigating damage from environmental stresses through dynamic shape transformations, including minimizing surface area for survival against desiccation and enabling transport from hazardous environments\cite{ozkan2021collective}. These examples show that physically interaction enables swarms to achieve challenging tasks that are not feasible when performed individually. {\WD Slight rephrasing of the previous sentence for better coherence.}

\begin{figure}[bhpt]
\centering
\includegraphics[width=8.7cm]{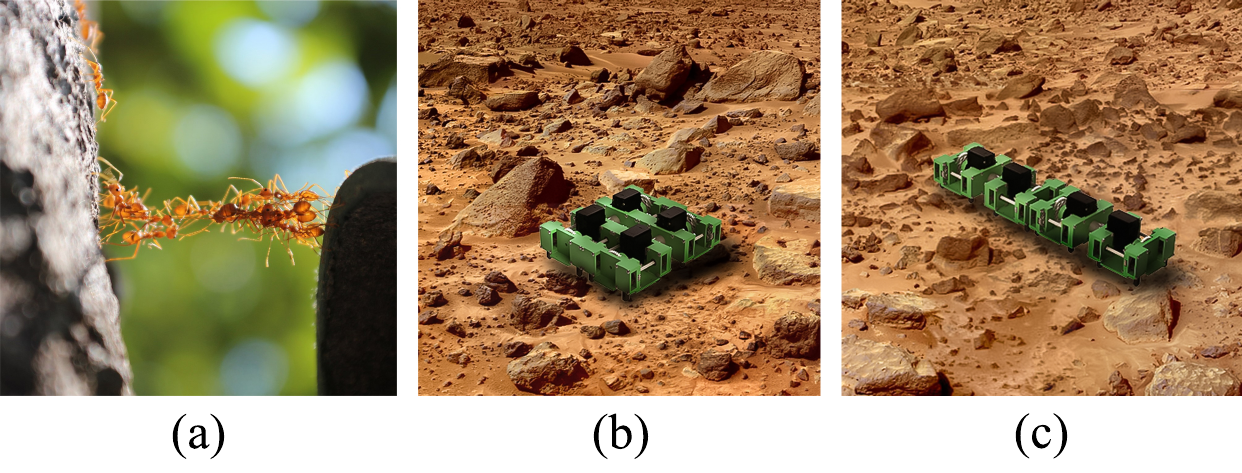}
\caption{Animals and robots can utilize physical connections to navigate challenging environments. (a) Ants collectively overcome a large gap by physically connecting with one another. (b, c) Multiple robots form different physical connection configurations to negotiate complex terrains.}
\label{fig:robot cooperate on challenging terrain}
\end{figure} 

Many physically-interacting swarms have been developed~\cite{davey2012emulating, romanishin2013m, boudet2021collections,chin2022flipper, saintyves2024self, rubenstein2012kilobot, chvykov2021low, zhou2021collective, zhou2020rapid} with capabilities of rearranging the configuration of its modules to achieve desired collective shapes (\eg lattice, chain, etc). One challenge is that swarm control often requires complex planning and control for each agent to achieve the collective behavior. To address this challenge, recent studies have investigated the mechanical interactions between swarm robots to enable highly-simplified collective control strategies~\cite{savoie2019robot,li2019particle,ozkan2021collective,saintyves2024self, goldman2024robot, li2021programming, ozkan2021self, zhou2021collective, chvykov2021low}. Two recent studies found that by loosely coupling with one another through statistical mechanics principles, a group of particle-like robots could generate desired collective trajectories without algorithmic control~\cite{li2019particle,savoie2019robot}. These studies demonstrated that with a better understanding of principles governing robot-robot interactions, a robot swarm could accomplish complex tasks without extensive controls and computation. 

Another challenge in extending the applicability of swarm systems to real-world scenarios is that currently most of these systems are demonstrated in relatively simple environments that are flat and rigid. To enable the next-generation swarm systems that can produce ants-like collective mobility and cope with challenging terrains, a better understanding of the robot-environment interactions is required. A recently-developed ``obstacle-aided locomotion'' framework~\cite{qian2019obstacle,chakraborty2022planning,Haodi2024obstacle, hulearning} represented the physical environments as ``interaction force opportunities'' and enabled simple robots to utilize environment interactions to effectively traverse extreme terrains. As a beginning step to extend this framework to multi-agent systems, our paper  %{\WD I suggest to change "systematically" to "thoroughly" to avoid sounding repetitive.} 
studied how different physical connections could allow a robot group to couple with terrain features and produce desired motion.

The major contributions of this work are summarized as follows:
\begin{itemize}
    \item Through systematic experiments, discovered that physical configuration in the multi-robot system can significantly affect their capability to collectively traverse densely distributed large obstacles that were inaccessible to individual agents. 
    \item Developed an energy-based model that can reveal the mechanism behind the observed effect, and predict configuration parameters that can enable collective traversal for a diverse range of terrain and robot parameters
    \item Demonstrated that the model-predicted configuration parameters can enable extremely simple control of multi-agent systems to collectively traverse challenging obstacle terrains
\end{itemize}
\section{Method}\label{sec:method}
%\colorbox{green}{Feifei +1} \colorbox{purple}{Haodi +3}

To explore the effect of physical connections on collective robot locomotion, we developed robots with limited individual locomotion ability %{\WD Suggest to change to "ability".} 
but are capable of connecting to one another to cope with rough terrains. We performed laboratory experiments to study how different connection length between the robots influences their collective traversability on rough terrains. 

% \begin{figure}[bhtp]
% \centering
% \includegraphics [width=8.8cm]{Figures/png/Fig-2.png}
% \caption{Robot mechanical design. (a) is a gear transition structure designed to drive individual legs, (b) is a peg leg along with a circular plate, (c) is a complete robot with one drive motor.}
% \label{fig:robot cad}
% \end{figure} 

\subsection{Robots}\label{sec:robot}

\begin{figure}[tbp]
\centering
\includegraphics[width=8.0cm]{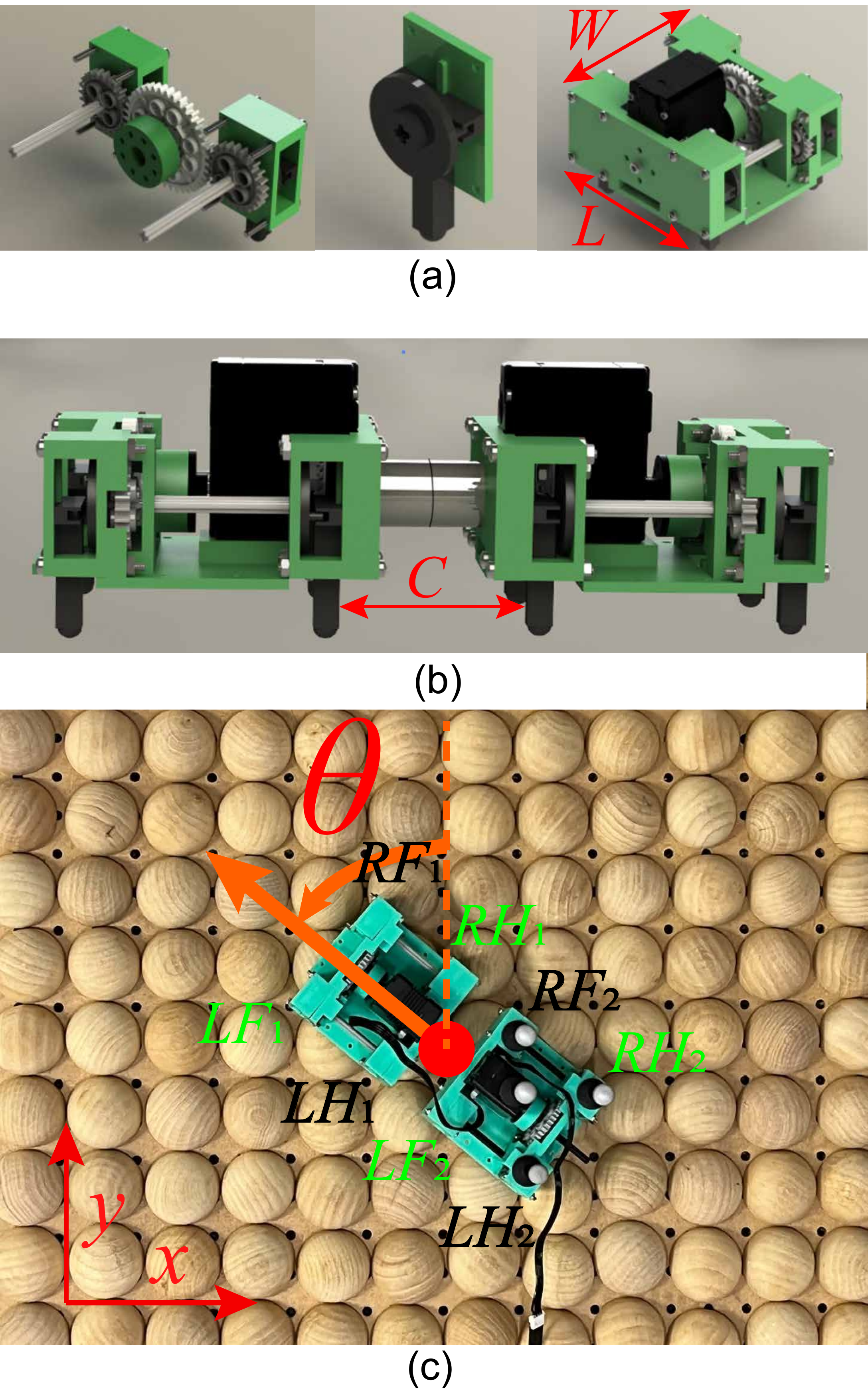}
\caption{Robot and Experiment setup. (a) The design of each individual robot. (b) A two-robot system connected to each other via an electrical magnet. (c) Locomotion experiment setup, where the two-robot system move across an bumpy terrain comprising of uniformly distributed semispherical boulders. $LF$, $RF$, $RH$, $LH$ denote the left front, right front, right hind, and left hind leg of each robot, respectively. Leg $LF_1$, $RH_1$, $LF_2$, $RH_2$ (``leg group 1'') move synchronously, and alternate with the other four legs, $RF_1$, $LH_1$, $RF_2$, $LH_2$ (``leg group 2''). $\theta$ denotes the orientation angle of the two-robot system in the yaw direction.}
\label{fig:obstacle environment}
\end{figure}

The body of each individual robot was 6.3cm $\times$ 6.3cm, 3D printed using PLA. %We defined the robot's size with body width, $W$, which is the distance between the center of two front robot legs, and body length $L$, which is the distance between the center of two side legs, and robot connection length $C$, which is the distance between the connected two robots. 
To better investigate how leg-obstacle interaction force could affect robot dynamics, we used 4 vertically peg legs for each robot to decouple the obstacle-induced robot displacement from the robot self-propulsion. Each leg's linear motion was achieved through a Scotch yoke mechanism, where a pin engaged a slot on a \SI{2.5}{\centi\meter} circular plate, producing a simple harmonic motion with a \SI{2.5}{\centi\meter} range. The modular leg design enabled flexible phase and gait adjustments. To synchronize leg movement, a gear system was introduced, with a main gear and two smaller side gears connected to leg pairs  (Fig.\ref{fig:obstacle environment}a). A gear housing facilitated gear support and leg attachment to the robot body, ensuring proper phase coordination through a single motor (Lynxmotion LSS-ST1) per robot. A trotting-like gait pattern was preset for each individual robot, where two diagonal legs (Fig. \ref{fig:obstacle environment}c, $LF_i, RH_i$) move synchronously and alternate with the other diagonal pair (Fig. \ref{fig:obstacle environment}c, $RF_i, LH_i$). Here $LF$, $LF$, $RH$, $LH$ represent the left front, right front, right hind, and left hind leg, respectively, and $i \in \{1, 2\}$ denotes the number of robots. For all results reported in this paper, the stride frequency was set to 0.33 Hz.

To focus on collective mobility, each robot was only equipped with vertical %{\WD Maybe add the word "vertical" here to make it easier for readers to understand why the robot can't move} 
actuation, and individually incapable of producing displacements, neither on rough terrain nor on flat ground. To investigate multi-robot dynamics on the obstacle terrains, we designed each individual robot to be connectable to one another Three connection mechanisms were implemented: (i) through an electrical magnet (Fielect FLT20190821M-0016, Fig. \ref{fig:obstacle environment}b); (ii) through a rigid connector with varying lengths; and (iii) through a mini linear actuator (DC House LA-T8-12-50-30/85-20). Connection mechanism (i) and (ii) were used in systematic locomotion experiments to study the effect of connection length on collective traversability, whereas connection mechanism (iii) was used in the demo experiments to illustrate the potential for the robots to flexibly adapt their connection length or direction for traversing different terrain features. To ensure that two-robot system are symmetric and its center-of-mass is situated in the middle of the system for all three connection mechanisms, the mounting locations of the inter-robot connectors are positioned at the midpoint between $LH_1$ and  $RH_1$ for robot 1, and the midpoint between $LF_2$ and  $RF_2$ for robot 2. %{To ensure that the overall footprint of two connected robots is identical regardless of the mounting direction of the second robot, for all three connection mechanisms the mounting locations for the inter-robot connectors were positioned on each side of the robot and in between two legs.} {\WD \{I suggest to change the curly-braced section before this to the following: \}To ensure that two-robot system are symmetric and its center-of-mass is situated in the middle of the system for all three connection mechanisms, the mounting locations of the inter-robot connectors are positioned at the midpoint between $LH_1$ and  $RH_1$ for robot 1, and the midpoint between $LF_2$ and  $RF_2$ for robot 2.} 
The connection length between the robots, $C$, can be systematically varied with connection mechanism (iii), to study how physical connection configuration influences robot-terrain coupling. In this study, we began with a two-robot system to study the effect of physical connection lengths. %We discuss in Sec. \ref{sec:conclusion} how our results can be extended to systems with larger number of robots. 

%We define robot state$(\bar X, \bar Y, \theta)$, as a combination of modulated CoM position on the x-axis, $\bar X = X |_{P_x}$, on the y-axis, $\bar Y = Y |_{P_y}$, and orientation, $\theta$. We used the modulated CoM position to represent the robot state because of the periodic distribution of obstacles. Since we implement trot gait with the same phase for the two robots, we divided the robot legs into two groups, $G_1$ and $G_2$. $G_1 = \{ LF_1, RH_1, LF_2, RH_2 \}$ and $G_2 = \{ RF_1, LH_1, RF_2, LH_2 \}$, where $LF$ represents left front leg, $RH$ represents right hind leg, $RF$ represents right front leg and $LH$ represents left hind leg, the index 1 and 2 are the front robot and the hind robot, respectively.

%In order to distinguish each individual robot in multi-robot we assigned a unique number $n$ to each individual robot. For example, a single robot is assigned number 1, and individual robots in a connected robot consist of two single robots that are assigned numbers 1 and 2.

\subsection{Rough Terrain Locomotion Experiments}
The locomotion experiments were performed on laboratory ``rough terrain'', consisting of a peg board with wooden semispherical ``boulders'' on top (Fig. \ref{fig:obstacle environment}c). The diameter of the boulders was chosen to be 5 cm to emulate challenging rough terrains where terrain height variation is comparable with the robot leg length. 
To create a challenging terrain with densely distributed boulders, the boulders were placed adjacent to one another without gaps along both $x$ and $y$ directions. To capture the collective dynamics of the robot group as they traverse the obstacle field, we used four cameras (Optitrack Prime 13W) at the four corners of the experiment arena to track robot states in the world frame, and two additional cameras (Optitrack Prime Color) to record experiment footage. The tracked state of the robot group includes the position of the geometric center of the connected robot pair, as well as the pitch, roll, yaw angles of the connected robots (Fig. \ref{fig:obstacle environment}). Both tracking data and video were recorded at a frame rate of 120 frames per second (FPS).

A total of 45 locomotion experiments were performed, with 15 systematically-varied robot connection lengths. We first tested 7 connection lengths of \SI{4.0}{\centi\meter} to \SI{7.0}{\centi\meter}, with an increment of \SI{0.5}{\centi\meter} connection length. We found that the connected robot exhibited a significant stride-wise displacement with the  \SI{5.5}{\centi\meter}, whereas all other connection lengths exhibited almost zero displacements. Intrigued by this finding, we selected 8 additional connection lengths around \SI{5.5}{\centi\meter}, from \SI{5.1}{\centi\meter} to \SI{5.4}{\centi\meter}, and \SI{5.6}{\centi\meter} to \SI{5.9}{\centi\meter}, both with an increment of \SI{0.1}{\centi\meter}. 3 trials were performed for each connection length. For all trials, we start the robot inside the obstacle fields with the same initial position and orientation.

\begin{comment}
The quadrupedal robot has a square footprint with a length and width of 6.3cm. The length is measured from the center of right front leg(RF) to the center of the right hind leg(RH). Similarly, the robot width is measured from the center of the left front leg(LF) to the center of the right front leg(RF). This square footprint allowed for simpler simulation and experimentation.

Two robots were connected together through the use of a 5.5 cm long connector; Measured from the edge of the footprint of the first robot to the edge of the  footprint of the adjacent robot. To ensure symmetry, the four mounting locations for the connector is positioned on each side of the robot and in between two legs. As a result the overall footprint of two connected robots is identical regardless of the mounting location of the second robot.

All four legs are driven by a single servo motor using three gears and two axles. To change the servo's rotational motion to linear motion, each leg implemented a Scotch Yoke mechanism. A  pin, on a 2.0 cm circular plate, engages a slot that drives the reciprocating leg. The displacement of the leg in respect to time follows simple harmonic motion. 

Robot weight is xxx g and implemented with trot gait, meaning the RF leg and LH leg are synchronous and alternate with LF leg and RH leg. This gate, along with the other dimensions, was based off of the simulation results that offered the largest displacement.
\end{comment}
\section{Results}\label{sec:results}

\subsection{Small changes in connection length can lead to significant differences in traversability}
%\colorbox{green}{Feifei +1} \colorbox{purple}{Haodi +2}

\begin{figure*}[htbp!]
    \centering
    \includegraphics[width=16cm]{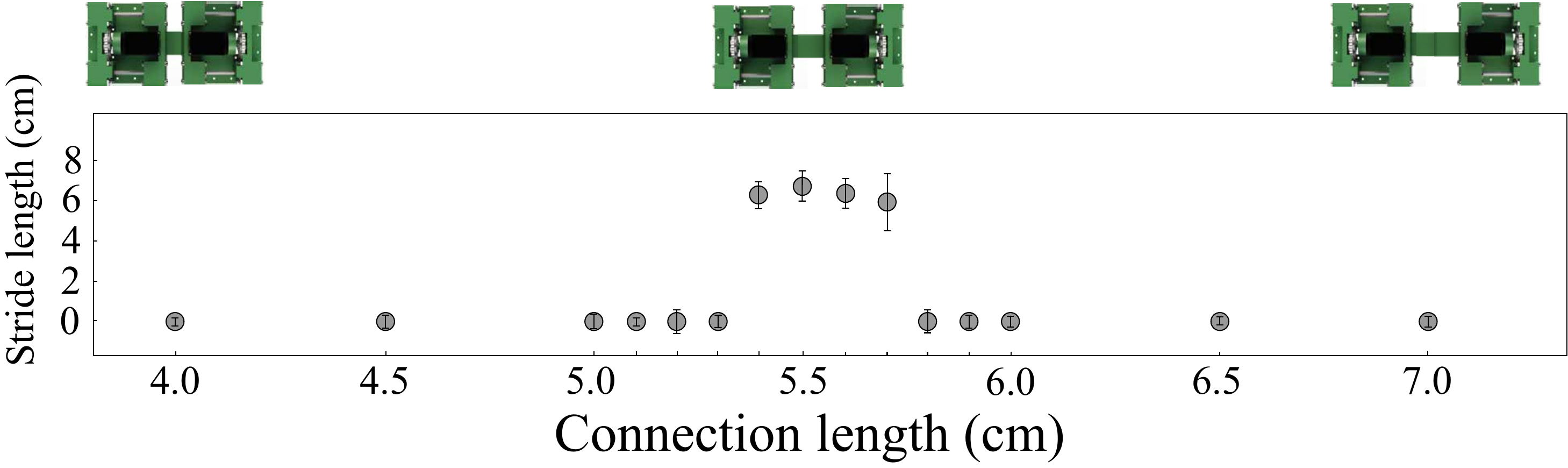}
    \caption{Experimentally-measured robot stride length (\ie displacement during each cycle) with different robot connection lengths. Markers represent robot stride-wise displacement averaged from all steps from the same connection length. Error bars represent one standard deviation.}
    \label{fig:multi-robot experiment for position and connection length}
\end{figure*}

We observed significant differences in robot speed across the obstacle field, as we systematically varied the connection length. With connection lengths $C$ between 5.4 cm (0.86 UBL) and 5.7 cm (0.90 UBL), the connected robots were found to produce significant displacements during each stride cycle (approximately one body length per cycle) across the obstacle fields (Fig.~\ref{fig:multi-robot experiment for position and connection length}). %, with the minimum being 6.1 $\pm$ 1.3 cm (with $C$ = 5.7 cm) and the maximum being 6.8 $\pm$ 0.8 cm (with $C$ = 5.5 cm) 
For connection lengths $C$ from 4 cm (0.63 UBL) to 5.3 cm (0.84 UBL), and from 5.8 cm (0.92 UBL) to 7.0 cm (1.1 UBL), however, the connected robot pair oscillated in place with approximately zero stride-wise displacements (Fig.~\ref{fig:multi-robot experiment for position and connection length}).

%We also noticed that depending on the initial position, the robot collectives would move towards one of the four directions: $\theta = 45^{\circ}$, $135^{\circ}$, $-135^{\circ}$, $-45^{\circ}$ (Fig.\ref{fig:multi-robot experiment for position and connection length}(b)). 

\subsection{Transition from collective flowing to collective jamming} \label{sec: jamming-states} %Robot exhibited an ``unjamming'' to ``jamming'' transition during each stride cycle %``Jamming'' states were observed at local minima of gravitational potential energy
%\colorbox{green}{Feifei +1} \colorbox{purple}{Haodi +2}

\begin{figure}[htbp!]
    \centering
    \includegraphics[width=8.8cm]{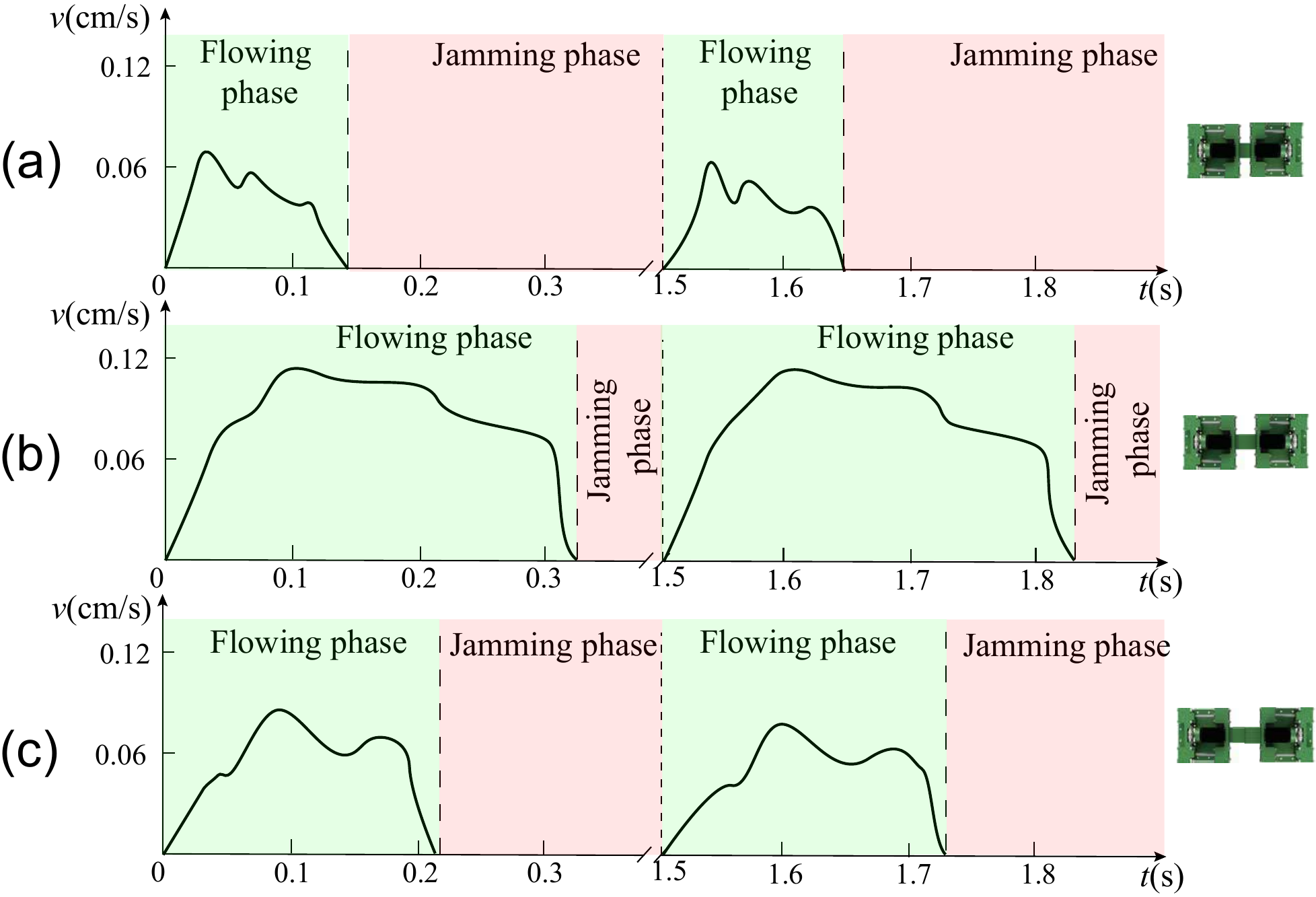}
    \caption{Experimentally-measured CoM velocity of the two-robot system during one stride period, for (a) connection lengths \SI{5.0}{\centi\meter}, (b) \SI{5.5}{\centi\meter}, and (c) \SI{6.0}{\centi\meter}. The red and green shaded color regions represent the collective flowing phase and collective jamming phase, respectively.}
    \label{fig:two-phases}
\end{figure}

To understand the difference between the two significantly different traversability groups, we analyze the robot center-of-mass (CoM) velocities within each stride cycle. The tracked robot velocities showed that regardless of robot connection length, the connected robots exhibited two phases during each stride:
(1) a ``flowing'' phase (Fig. \ref{fig:two-phases}, Green phase), whereupon the touchdown of robot legs, the robot pair generates a large velocity on the horizontal plane under the leg-obstacle interaction forces.
(2) a ``jamming'' phase (Fig. \ref{fig:two-phases}, Red phase), where the robot CoM velocity in the world frame decreases to zero.
%robot legs slide from the obstacle surface and leg-obstacle interaction forces either become 0 or cancel each other which results in the robot keeping still.

Interestingly, while both phases were observed from all connection lengths, the connection length were found to affect the duration of the flowing phase. The duration of the flowing phase was found to be significantly longer with the connection lengths that exhibited the ``traversal'' behavior (\eg \SI{5.5}{\centi\meter}, Fig. \ref{fig:two-phases}b), as compared to those that exhibited the ``stuck'' behavior (\eg \SI{5.0}{\centi\meter} and \SI{6.0}{\centi\meter}, Fig. \ref{fig:two-phases}a, c).

\subsection{Collective traversability is governed by the direction of transition between jamming states}\label{sec:mechanism1}
%\colorbox{purple}{Haodi +2}

To understand how connection length affects the duration of the ``flowing phase'', we investigated two questions in this section: (i) what is the condition for robots to switch from collective flowing to collective jamming? (ii) how is the duration of the flowing phase related to the jamming condition? 

\begin{figure}[thbp]
    \centering
    \includegraphics[width=8.8cm]{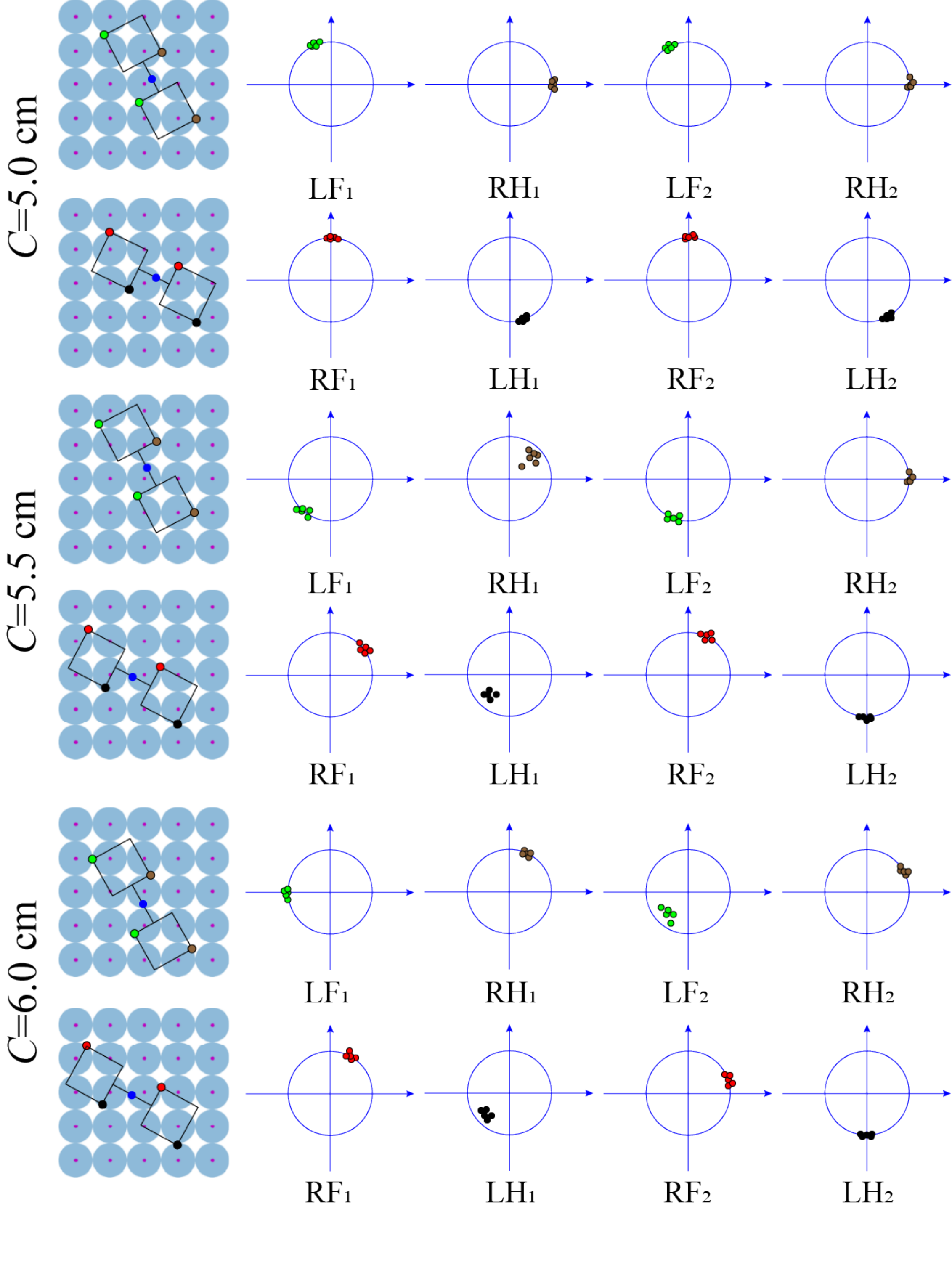}
    \caption{Experimentally-measured robot legs contact positions when the connected robot were during the jamming phase, for 3 representative connection lengths: (top) $C$ = \SI{5.0}{\centi\meter}, (middle) \SI{5.5}{\centi\meter}, and (bottom) \SI{6.0}{\centi\meter}. The robot states in the left diagram illustrated the experimentally-observed robot jamming states. Red, black, green, brown markers represent the contact positions of $LF$, $RH$, $RF$, $LH$ legs on the semispherical boulder (represented as blue circles) from top view. The leg contact positions plotted were measured from the last 5 strides of the 3 trials for each connection length.}
    \label{fig:robot legs contact positions}
\end{figure}

To determine the jamming condition, we analyzed the experimentally-tracked leg-terrain contact position when the connected robot shifted from the flowing phase to the jamming phase. We found that during the jamming phase, all robot legs were located within proximity ($\leq$ \SI{0.8}{\centi\meter}) of the obstacle edge (Fig.~\ref{fig:robot legs contact positions}). We hypothesized that it was governed by the robot-terrain coupling. Due to the relatively low frictional coefficient ($\mu=0.08$) between robot legs and obstacles, the robot state (position and orientation) within the world frame was primarily driven by the gravitational forces. Depending on the connection length, the connected robots would couple with the same terrain differently and reach the lowest potential energy state at different positions (which we refers to as the ``jamming state'' in the remaining of the paper). These jamming states thus determine the switching condition for the robots to shift from the flowing phase to the jamming phase.

\begin{figure}[hbtp!]
    \centering
    \includegraphics[width=0.95\linewidth]{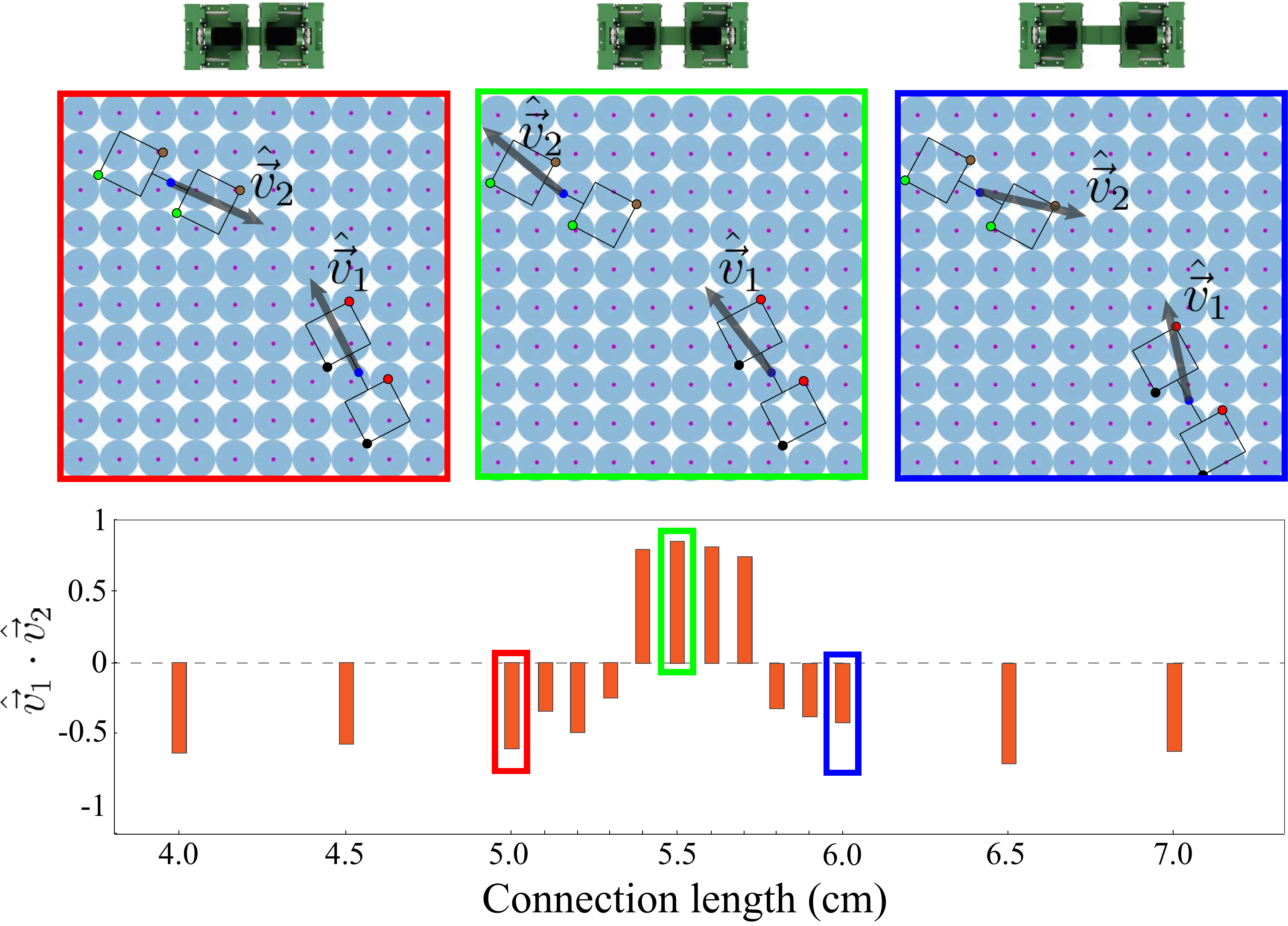}
    \caption{Experiment measured $\hat{\vec{v}}_1 \cdot \hat{\vec{v}}_2$. Diagrams highlighted in red, green, and blue boxes illustrated the robot collective flowing directions, $\hat{\vec{v}}_1$ and $\hat{\vec{v}}_2$ for connection length \SI{5.0}{\centi\meter}, \SI{5.5}{\centi\meter}, \SI{6.0}{\centi\meter}, respectively. The bottom plot shows the experimentally-measured $\hat{\vec{v}}_1 \cdot \hat{\vec{v}}_2$ for all connection lengths tested.}
    \label{fig:experiment vector product}
\end{figure}

To relate the duration of flowing phase to the jamming condition, we analyzed the instantaneous robot velocity at the beginning of each collective flowing phase, $\vec{v}_1$, $\vec{v}_2$. Here $\vec{v}_1$ and $\vec{v}_2$ represent the robot pair's velocity shortly after the leg group 2 and 1 touchdown, respectively. Interestingly, we noticed the sign of $\hat{\vec{v}}_1 \cdot \hat{\vec{v}}_2$ exhibited a high correlation with the robot's flowing phase duration and collective mobility (Fig. \ref{fig:experiment vector product}). For the range of connection lengths where the connected robots exhibited long flowing phase and high traversability, the dot product between the $\hat{\vec{v}}_1$ and $\hat{\vec{v}}_2$ is positive (Fig. \ref{fig:experiment vector product}, 5.4cm - 5.7cm connection length). On the other hand, for the range of connection lengths where the connected robots exhibited short flowing phase and low traversability, the dot product between the $\hat{\vec{v}}_1$ and $\hat{\vec{v}}_2$ is negative (Fig. \ref{fig:experiment vector product}, connection length between 4 - 5.3 cm, and 5.8 - 7.0 cm). This can be understood intuitively: if the dot product of the robot's speed during both steps within one stride was positive, the robots moved towards similar direction from the two steps, and the two step lengths produce a ``constructive'' effect (\ie added together), resulting in high mobility; if the dot product of the robot's speed during both steps within one stride was negative, the robots move towards opposite direction from the two steps, and the two step lengths produce a ``destructive'' effect (\ie canceling each other out), resulting in low mobility. 

This finding explained the observed differences in traversability for the two-robot system. Next, we investigate how the sign of $\hat{\vec{v}}_1 \cdot \hat{\vec{v}}_2$ was modulated by the robot connection length (Sec. \ref{sec:mechanism2}).

%(move to Sec. V) As the robot's position at the beginning of each leg pair's active phase is the same with the other leg pair's jamming state, the legs' contact positions can be inferred. 

% \subsection{acceleration speed for robot with different connection lengths}

% \begin{itemize}
%     \item discuss the observation that for all connection lengths, there are two primary phases upon touchdown: (i) flowing/advancing (describe the behavior); (ii) jamming (describe the behavior)
%     \item discuss the finding that the ratio of flowing (unjammed) duration and jammed duration differs significantly for \SI{5.5}{\centi\meter}, \SI{5.0}{\centi\meter}, and \SI{6.0}{\centi\meter}. With the connection length where large robot speed was observed (\SI{5.5}{\centi\meter}), the ``duty factor'' of unjammed duration was significantly larger as compared to connection length were small robot speed was observed (\SI{5.0}{\centi\meter}, \SI{6.0}{\centi\meter}).
% \end{itemize}

\subsection{An energy landscape based model explains the modulation of collective traversal directions via connection length}\label{sec:mechanism2}

%\colorbox{purple}{Haodi +1}\colorbox{green}{Feifei +1} \colorbox{yellow}{Xingjue +1}

To investigate how the connection length modulates the collective flowing directions, we adopted an energy landscape~\cite{othayoth2020energy} framework to compute the system energy for a given connection length and obstacle spacing. The energy landscape framework assumes that the robot state will converge towards the lowest system energy state, and it uses measured physical and geometric parameters of the robot and the environment to compute the robot's system potential energy (gravitational and elastic) as a function of body rotation. %{\XL technically the model also needs the position of RH2/LH2} 
Despite its highly-simplified assumptions, this framework has shown great success in capturing dominating attracting states during locomotor-environment interactions~\cite{li2015cockroachgrass,han2021shape,othayoth2020energy}.

According to the energy landscape framework, the system's potential energy $E(X, Y, Z, \alpha, \beta, \theta)$ is a function of the robot state, where $X, Y, Z$ represent the CoM position of the connected robots in the world frame, and $\alpha, \beta, \theta$ represent the pitch, roll, yaw of the connected robots. By minimizing the system energy over body rotation, energy landscape at each state can be identified, which determines the robot transition direction~\cite{othayoth2020energy}. For our system, body rolling was small ($\beta =$ $7.67^{\circ} \pm 3.54^{\circ}$ across all trials), so for a given robot state ($X$, $Y$, $\theta$), we minimized the system's potential energy $E = mgZ$ over the body pitch $\alpha$, by allowing the robot body to freely pitch while finding the minimal $Z$ such that the robot legs do not penetrate through the ground or the obstacles. 

\begin{figure}[hbtp]
    \centering
    \includegraphics[width=1.0\linewidth]{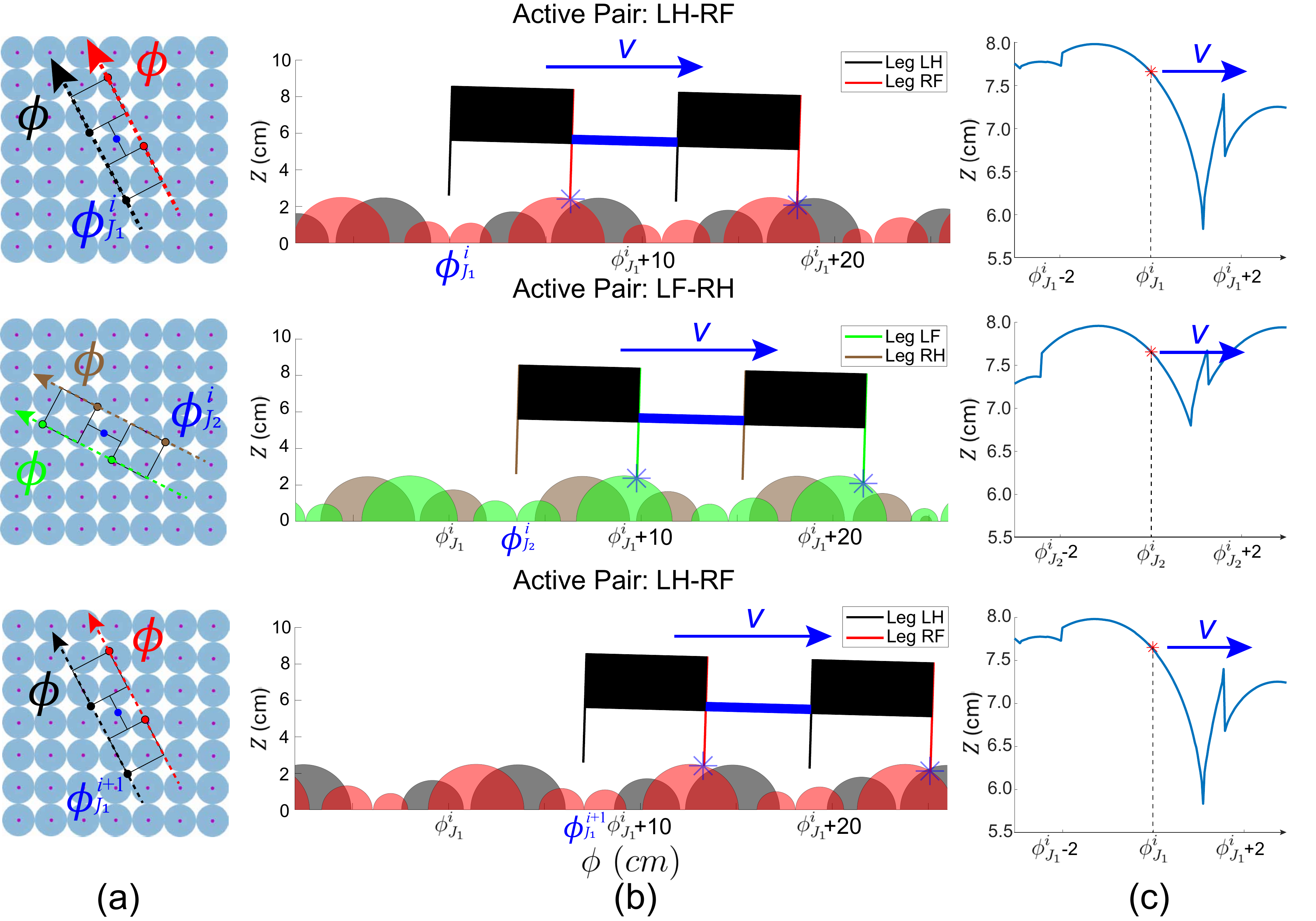}
    \caption{Energy model prediction for connection length \SI{5.5}{\centi\meter}. (a) Diagrams illustrating the robot state when transitioning from the jamming phase to the flowing phase. (b) Diagrams illustrating the saggital-plane robot leg contact pattern with projected obstacles on $\phi$ axes in (a). Green, brown, red, and black shaded semicircles represent the projected obstacles on the $\phi$ axes corresponding to the $LF$, $RH$, $RF$, $LH$ legs, respectively. $\phi_{J_1}^i$ and $\phi_{J_2}^i$ denote the position of $LH_2$ and $RH_2$ legs at the step $i$ jamming state. (c) Energy landscape of the robot CoM. Blue arrows in (b) and (c) represent robot collective flowing directions.}
    \label{fig:55mm}
\end{figure}

\begin{figure}[hbtp]
    \centering
    \includegraphics[width=1.0\linewidth]{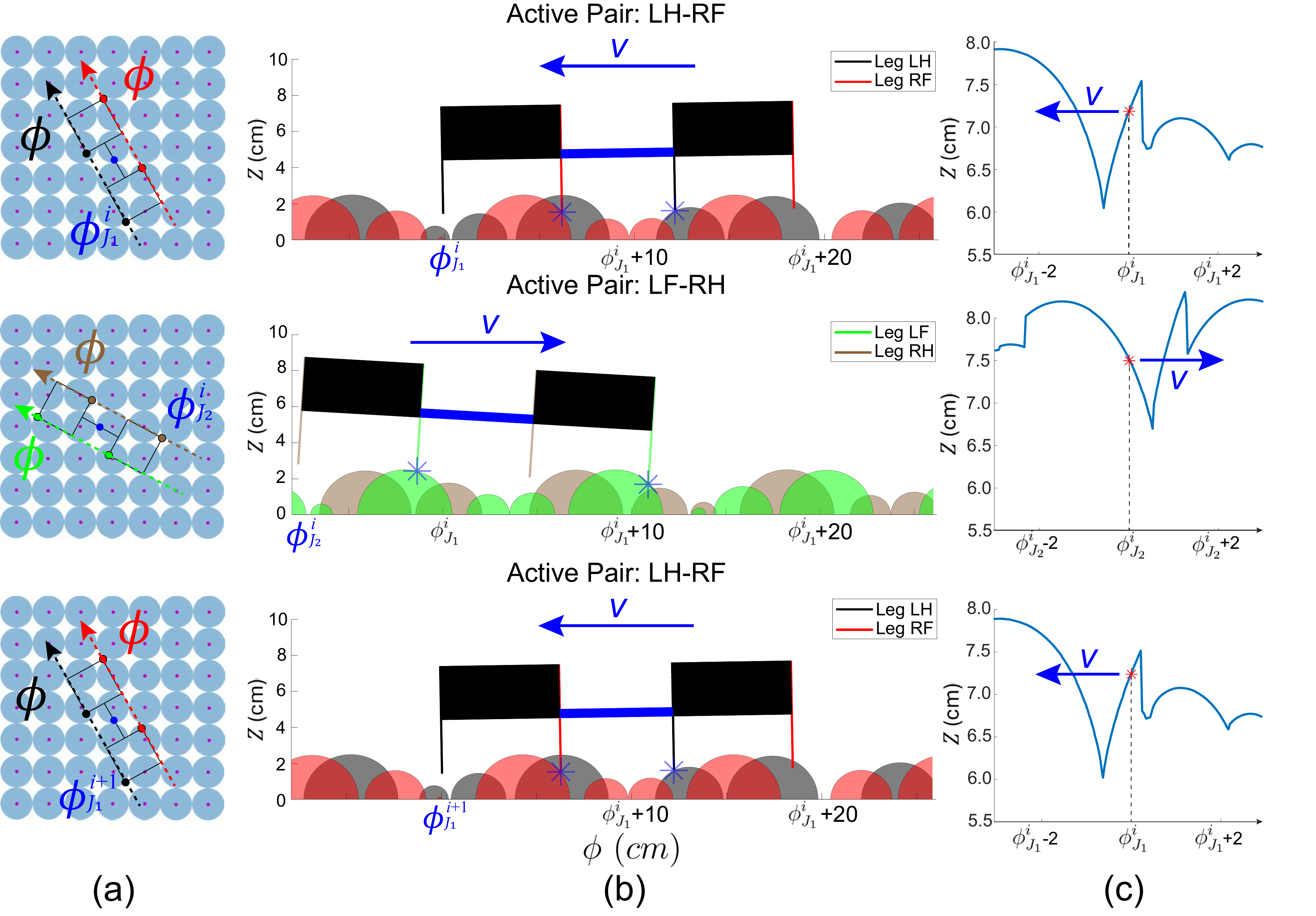}
    \caption{Energy model prediction for connection length \SI{6.0}{\centi\meter}. (a) Diagrams illustrating the robot state when transitioning from the jamming phase to the flowing phase. (b) Diagrams illustrating the saggital-plane robot leg contact pattern with projected obstacles on $\phi$ axes in (a). Green, brown, red, and black shaded semicircles represent the projected obstacles on the $\phi$ axes corresponding to the $LF$, $RH$, $RF$, $LH$ legs, respectively. $\phi_{J_1}^i$ and $\phi_{J_2}^i$ denote the position of $LH_2$ and $RH_2$ legs at the step $i$ jamming state. (c) Energy landscape of the robot CoM. Blue arrows in (b) and (c) represent robot collective flowing directions.}
    \label{fig:60mm}
\end{figure}

To determine the direction of collective flowing starting from the previous step's jamming phase, we computed the energy landscape at the two jamming states, ($X_{J1}, Y_{J1}, \theta_{J1}$) and ($X_{J2}, Y_{J2}, \theta_{J2}$). Fig.~\ref{fig:55mm}c and Fig.~\ref{fig:60mm}c visualize the minimized system energy (the ``energy landscape'') in the robot's sagittal plane (Fig.~\ref{fig:55mm}b, Fig.~\ref{fig:60mm}b), at the beginning of each flowing phase (\ie the jamming states from the previous step). This energy landscape allows determining the flowing direction of the connected robot system (Fig.~\ref{fig:55mm}b and Fig.~\ref{fig:60mm}b, blue arrows) at those initial states, as the robot state would always converge towards the lower energy state (Fig.~\ref{fig:55mm}c, Fig.~\ref{fig:60mm}c, blue arrows).  

The energy landscapes explained how the connection length modulated the robots' collective traversability. For connection length between \SI{5.4}{\centi\meter} and \SI{5.7}{\centi\meter} (Fig.~\ref{fig:55mm}) which exhibited a high traversability across the bumpy terrains in experiments, the energy landscape at the beginning of two flowing phases predicted moving directions towards the same direction (Fig.\ref{fig:55mm}b, blue arrows), producing a ``constructive'' displacement across the bumpy terrain. As the robot connection length shortens or lengthens, the contact positions of the robot legs on the bumpy terrain gradually varies, resulting in a shifted energy landscape (Fig.~\ref{fig:55mm}c, Fig.~\ref{fig:60mm}c, position of red marker relative to the blue energy landscape). As a result, when connection length reaches above \SI{5.8}{\centi\meter} or below \SI{5.4}{\centi\meter} (Fig.~\ref{fig:60mm}), the energy landscape at the beginning of two flowing phases predicted moving directions towards the opposite direction (Fig.\ref{fig:60mm}b, blue arrows), producing a ``desctructive'' displacement that cancels each other out, leaving the robot oscillating in place. These model-predicted robot moving directions and resulting traversability agreed well with the experimental measurements (Fig. \ref{fig:experiment vector product}, Fig. \ref{fig:multi-robot experiment for position and connection length}). In Sec. \ref{sec:application}, we show that the model can also guide multi-robot connection length adaptations to enable collective traversal of bumpy terrains with varying spatial densities.

\section{Model-predicted connection length adaptation to enable collective traversal across challenging terrains}\label{sec:application}

%\colorbox{purple}{Haodi +2}\colorbox{green}{Feifei +1}

In this section, we show that the proposed energy landscape based model could help determine robot connection length adaptation for collectively traversing bumpy terrains with varying spatial frequencies. We follow a two-step procedure to achieve this: (i) identifying jamming states; and (ii) determine collective traversability.

To identify the jamming states, we computed the total leg-obstacle interaction forces following methods from ~\cite{qian2019obstacle}. By identifying all states where zero leg-obstacle interaction forces are zero, we have a small subset of potential jamming states. From those, we further selected states that map to themselves after a full stride cycle following the method from~\cite{Haodi2024obstacle} to identify the jamming states.

For determining collective traversability, we computed the energy landscape (Fig. \ref{fig:55mm}, Fig. \ref{fig:60mm}) for each feasible connection length, and checked the velocity vector direction at each jamming state. If the product of the velocity vector of the given connection length is negative then we pick up another connection length and repeat the previous step until finding a connection length that has a velocity vector production value larger than 0. We then use this connection length for the robot to traverse the corresponding obstacle field. 

\begin{figure}[bhtp!]
    \centering
    \includegraphics[width=0.95\linewidth]{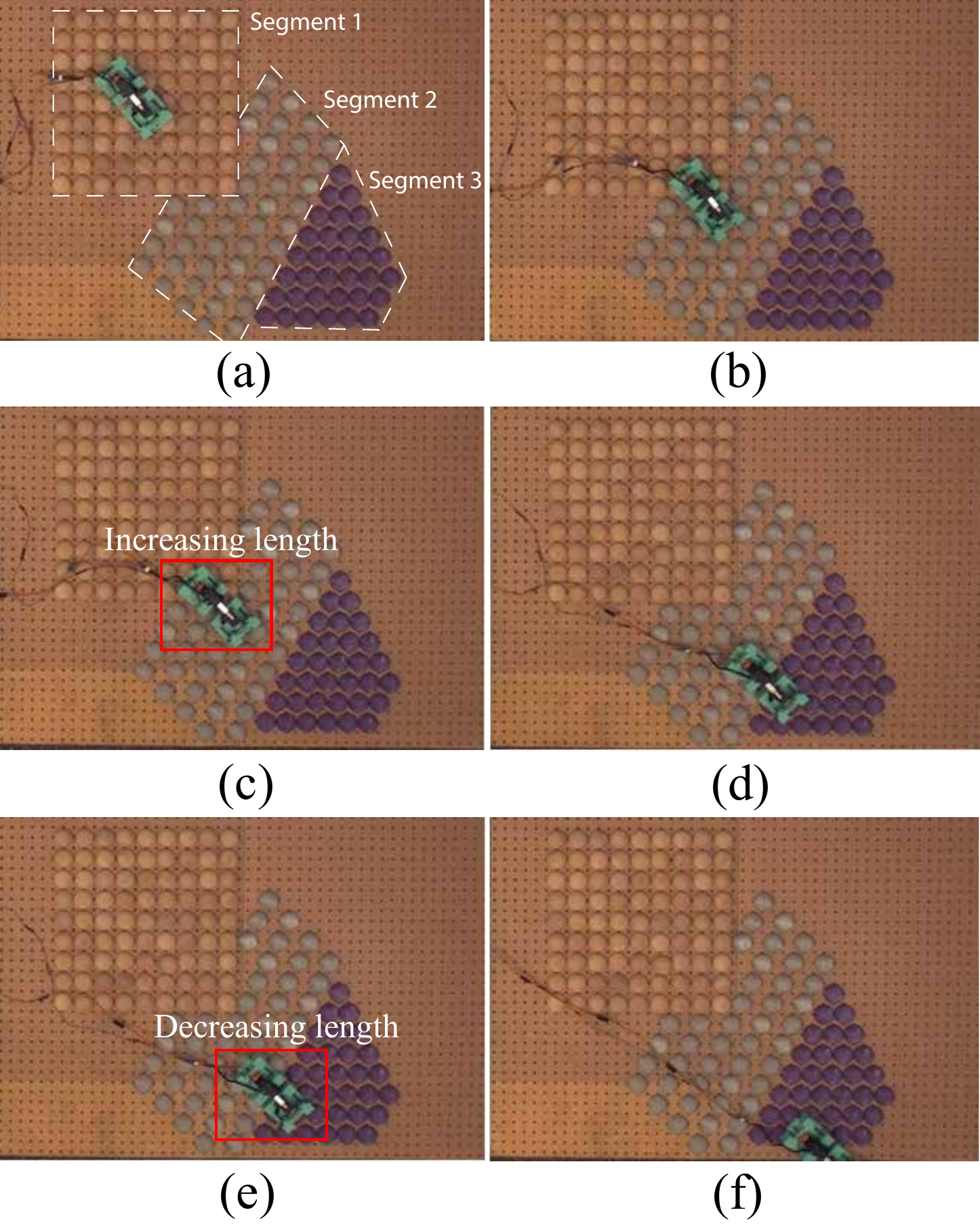}
    \caption{Robot traversal different spacing obstacles fields. Yellow, gray, and purple obstacle fields are 3 obstacle fields with different spacing. (a) is the robot start in the yellow obstacle fields; (b) is the robot move from its starting position to the intersection area between yellow and gray obstacle fields and get stuck; (c) is the robot change its connection length and continue to move to the intersection area of gray and purple obstacles fields and is stuck at (d); (e) is the robot changes its connection length and continue to move to the obstacle fields boundary at (f).}
    \label{fig:adjusting connection length}
\end{figure}

To test the feasibility of our method, we challenged the two-robot system to collectively traverse a 3-segment obstacle field (Fig. \ref{fig:adjusting connection length}) by adapting the connection length between the robots. The 3 segments were set up with different obstacle density: segment 1 (Fig. \ref{fig:adjusting connection length}, yellow obstacle region) was set to be the same obstacle density as in our locomotion experiments (Sec. \ref{sec:method}); segment 2 (Fig. \ref{fig:adjusting connection length}, gray obstacle region) with more sparse obstacle density relative to segment 1; and segment 3 (Fig. \ref{fig:adjusting connection length}, purple obstacle region) with more dense obstacle distribution. For trials without connection length adaptation, the robots were observed to consistently get stuck at the boundary between the segments. 

Fig. \ref{fig:adjusting connection length} shows the experimental image sequence of the robot group traversing the 3-segment obstacle fields with model-informed connection length adaptation. The robot started in segment 1, with a connection length of $C_1$ = 5.5cm (Fig. \ref{fig:adjusting connection length}a). As the robots reached the intersection area between segments 1 and 2 (Fig. \ref{fig:adjusting connection length}b), the robot switches to $C_2$ = 7.0cm (Fig. \ref{fig:adjusting connection length}c), which is the model-informed connection length for the obstacle spacing in segment 2. Similarly, the robots switched to model-informed connection length for segment 3, $C_3$ = 4.5 cm, as it reached the intersection area between segments 2 and 3 (Fig. \ref{fig:adjusting connection length}d, e), and moved towards the desired destination on the bottom right (Fig. \ref{fig:adjusting connection length}f).

% \subsection{Two-robot system: dependence of moving direction on neighboring-agent location before connecting}
% In Sec.\ref{sec:experiment}-C we found different robot initial states could affect robot translational movement. In order to understand how initial robot states result in different robot moving directions, we run a numerical simulation that tries every possible initial state in the robot state space and records the state transition path for every initial state to generate a basin of attraction map, Fig.{\HH basin map to show with a different configuration, the robot states jump from one basin to another basin.} The basin of attraction map reveals that by reconfiguring the robot, the robot state will jump to another state in the state space and thus result in the robot being attracted to different basins and showing movement in different directions.

\section{Conclusion}\label{sec:conclusion}
%\colorbox{purple}{Haodi +1}\colorbox{green}{Feifei +1}

In this paper, we investigated how collective traversability of a two-robot system was affected by their physical connection configurations. We found that with different connection length, the two-robot system could produce either collective advancing or collective jamming when coupled with uneven terrain features. Through a energy landscape based model, we revealed how the collective traversability was governed by the robot-terrain coupling, and modulated through inter-robot connection. This understanding allowed the two-robot system to use an extremely simple control to adapt their connections and successfully move across different uneven terrain features. The results from this study opens up new avenues for a group of connected robots to collectively negotiate challenging terrains by adapting their physical connection configurations. Future work can build upon these results and extend the strategy to a larger number of connected robots. These understandings can enable future swarms toward life-like collective intelligence to operate in diverse environments.

\section*{Acknowledgments}
This work is supported by funding from the National Science Foundation (NSF) CAREER award \#2240075, the NASA Planetary Science and Technology Through Analog Research (PSTAR) program, Award \# 80NSSC22K1313, and the NASA Lunar Surface Technology Research (LuSTR) program, Award \# 80NSSC24K0127. The authors would like to thank Elliot Meeks and Seojon Kwon for assisting with robot design and experiment setup. %Simon To for help with preliminary experiments.

\clearpage
\bibliographystyle{IEEEtran}
\bibliography{ref.bib}

% \newpage

% If you have an EPS/PDF photo (graphics package needed), extra braces are

% Use $\backslash${\tt{begin\{IEEEbiographynophoto\}}} and the author's name as the argument followed by the biography text.
% \vfill
\end{document}